%% file: main-2563-Fabbri-finalversion.tex
\documentclass[11pt,a4paper]{article}
\usepackage{times,latexsym}
\usepackage{url}
\usepackage{subcaption}
\usepackage{tabularx}
\usepackage{multirow}
\usepackage{makecell}
 \usepackage{multirow}
\usepackage[T1]{fontenc}
\usepackage{stfloats}

\usepackage[acceptedWithA]{tacl2018v2}
\usepackage{xspace,mfirstuc,tabulary}

\usepackage{adjustbox}

\newif\iftaclinstructions
\taclinstructionsfalse % AUTHORS: do NOT set this to true
\iftaclinstructions
\renewcommand{\confidential}{}
\renewcommand{\anonsubtext}{(No author info supplied here, for consistency with
TACL-submission anonymization requirements)}
\newcommand{\instr}
\fi

\iftaclpubformat % this "if" is set by the choice of options

\else

\fi
\newcommand*{\affaddr}[1]{#1}
\newcommand*{\affmark}[1][*]{\textsuperscript{#1}}

\include{helpers}

\include{helpers-new-tabs-figs}

\title{SummEval: Re-evaluating Summarization Evaluation}

\author{
        \textbf{Alexander R. Fabbri}\affmark[$\dagger$]~\Thanks{Equal contributions from authors}
  \quad \textbf{Wojciech Kry\'sci\'nski}\affmark[$\ddagger$]~\textsuperscript{$*$}\\
  \quad \textbf{Bryan McCann}\affmark[$\ddagger$]
  \quad \textbf{Caiming Xiong}\affmark[$\ddagger$]
  \quad \textbf{Richard Socher}\affmark[$\ddagger$] 
  \quad \textbf{Dragomir Radev}\affmark[$\dagger \ddagger$] \\
  \affaddr{\affmark[$\dagger$] Yale University} 
  \quad 
  \affaddr{\affmark[$\ddagger$] Salesforce Research} \\
  \{alexander.fabbri,dragomir.radev\}@yale.edu,  \\ \{kryscinski,bmccann,cxiong,rsocher\}@salesforce.com
}

\date{}

\begin{document}
\maketitle
\input{0-abstract}
\input{1-introduction}
\input{2-related-work}
\input{3-metrics-models}
\input{4-methodology}
\input{5-metric-evaluation}
\input{6-model-evaluation}
\input{7-conclusions}
\input{8-acknowledgements}
% \clearpage
\bibliography{tacl2020}
% \clearpage
\bibliographystyle{acl_natbib}
\clearpage
\input{9-appendix}
\end{document}

%% file: helpers.tex
\newcommand{\AnnotationExamplesTable}{
\begin{table*}[ht]
\begin{subtable}{\textwidth}
\centering
 \begin{adjustbox}{max width=\textwidth}
 \small
  \begin{tabular}{p{0.8\linewidth}|c|c}
    \textbf{Generated Summaries} & \makecell{Expert \\ scores (avg.)} & \makecell{Crowd-worker \\ scores (avg.)}\\ \hline 
  
 the queen's guard was left red-faced after he slipped on a manhole cover he lost his footing and slid sideways, knocking his bearskin on the side .  the embarrassed soldier quickly scrambled to his feet as his colleagues marched past as if nothing had happened .  tourist david meadwell recorded the unscheduled manouevre outside buckingham palace on thursday afternoon .  & \makecell[tc]{Coh: 5.0 \\ Con: 5.0 \\ Flu: 5.0 \\ Rel: 5.0} &  \makecell[tc]{Coh: 3.4 \\ Con: 3.8 \\ Flu: 3.4 \\ Rel: 3.8} \\ 
 
 \hline 
  holidaymaker david meadwell recorded the unscheduled manouevre outside buckingham palace . \textcolor{red}{he} lost his footing and slid sideways , knocking bearskin on the side of the box . queen 's guard was left red-faced after he slipped on manhole cover . \textcolor{red}{the entire incident was caught on a manhole cover }. the embarrassed soldier quickly scrambled to his feet as his colleagues marched past . & \makecell[tc]{Coh: 2.7 \\ Con: 2.0 \\ Flu: 4.7 \\ Rel: 3.7} & \makecell[tc]{Coh: 3.2 \\ Con: 3.4 \\ Flu: 3.4 \\ Rel: 4.0} \\ 
  \hline
  buckingham palace guard slipped on manhole cover in front of hundreds of horrified tourists . the queen 's guard was left red-faced after he slipped  on a manhole cover . he lost his footing and \textcolor{red}{dropped his rifle} on the side of the box and \textcolor{red}{dropping his rifle} . the incident was caught on \textcolor{red}{camera camera camera }.  the guard is thought to have slipped because of metal shutters nailed to the soles of his boots . & \makecell[tc]{Coh: 3.3 \\ Con: 5.0 \\ Flu: 1.7 \\ Rel: 4.3} & \makecell[tc]{Coh: 3.0 \\ Con: 3.2 \\ Flu: 2.8 \\ Rel: 3.2} \\
  \hline
  \end{tabular}
\end{adjustbox}
\caption{Generated summary examples illustrate common problems found in model outputs, such as ambiguous pronouns, incorrect references, and repetitive content.}
\label{annotation_divergence_a}
\end{subtable}
  
\begin{subtable}{\textwidth}
\centering
 \begin{adjustbox}{max width=\textwidth}
 \small
  \begin{tabular}{p{0.8\linewidth}|c|c}
  \textbf{Reference Summaries} & \makecell{Expert \\ scores (avg.)} & \makecell{Crowd-worker \\ scores (avg.)}\\ \hline 
  river plate admit they ` dream ' of manchester united striker radamel falcao . the colombia international spent eight years with the argentine club . falcao has managed just four goals in 19 premier league appearances . \textcolor{red}{read : falcao still ` has faith ' that he could continue at man utd next season . click here for the latest manchester united news .}  & \makecell[tc]{Coh: 3.0 \\ Con: 2.0 \\ Flu: 5.0 \\ Rel: 2.3} &  \makecell[tc]{Coh: 3.0 \\ Con: 3.6 \\ Flu: 3.0 \\ Rel: 4.4} \\ \hline
  \textcolor{red}{the incident occurred on april 7 north of poland in the baltic sea . u.s. says plane was in international airspace . russia says it had transponder turned off and was flying toward russia} & \makecell[tc]{Coh: 2.0 \\ Con: 1.7 \\ Flu: 3.0 \\ Rel: 2.3} &  \makecell[tc]{Coh: 4.0 \\ Con: 3.4 \\ Flu: 4.2 \\ Rel: 3.6} \\ \hline
  \end{tabular}
  \end{adjustbox}
  \caption{Reference summaries highlight issues found in the CNN/DailyMail dataset, such as click-baits and references to other articles as well as unreferenced dates and low coherence caused by concatenating bullet-point summaries.}
  \label{annotation_divergence_b}
  \end{subtable}
  \caption{Example summaries with the corresponding averaged expert and crowd-sourced annotations for \textit{coherence}, \textit{consistency}, \textit{fluency}, and \textit{relevance}.
  Expert annotations better differentiate coherence, consistency, and fluency among the examples when compared to the crowd-sourced annotations.}
   \label{annotation_divergence}
\end{table*}
}

\newcommand{\ModelsHumanJudgementScoreTable}{
\begin{table}[t]
\centering
\small
 \begin{adjustbox}{max width=\linewidth}
  \begin{tabular}{l|c|c|c|c}
    \textbf{Method} & Coherence & Consistency & Fluency & Relevance \\ \hline
   CNN/DM Reference Summary & 3.26 & 4.47 & 4.79 & 3.77 \\ \hline
    \multicolumn{5}{c}{\textit{Extractive Models}} \\ \hline
    M0 -  \textbf{LEAD-3} & \textbf{4.16} & 4.98 & 4.94 & 4.14 \\ \hline
    M1 -  \textbf{NEUSUM} & 3.22 & 4.98 & 4.90 & 3.82  \\ \hline % \citep{zhou-etal-2018-neural}
    M2 -  \textbf{BanditSum} & 3.28 & 4.99 & 4.83 & 3.81   \\ \hline % \citep{dong-etal-2018-banditsum} 
    M5 -  \textbf{RNES} & 3.71 & \textbf{4.97} & 4.81 & 4.06 \\ \hline % \citep{wu2018learning} 
    \multicolumn{5}{c}{\textit{Abstractive Models}} \\ \hline
    M8 -  \textbf{Pointer Generator} & 3.29 & 4.65 & 4.79 & 3.55 \\ \hline % \citep{see-etal-2017-get} 
    M9 -  \textbf{Fast-abs-rl} & 2.38 & 4.67 & 4.50 & 3.52   \\ \hline %  \citep{chen-bansal-2018-fast}
    M10 - \textbf{Bottom-Up} & 2.73 & 4.25 & 4.42 & 3.38   \\ \hline % \citep{gehrmann-etal-2018-bottom} 
    M11 - \textbf{Improve-abs} & 2.28 & 3.27 & 3.65 & 3.15   \\ \hline %  \citep{kryscinski-etal-2018-improving}
    M12 - \textbf{Unified-ext-abs} & 3.60 & \textbf{4.96} & 4.85 & 3.85 \\ \hline %  \citep{hsu-etal-2018-unified}
    M13 - \textbf{ROUGESal} & 3.44 & 4.82 & 4.86 & 3.83    \\ \hline %  \citep{pasunuru-bansal-2018-multi}
    M14 - \textbf{Multi-task (Ent + QG )} & 3.20 & 4.90 & 4.74 & 3.63 \\ \hline %  \citep{guo-etal-2018-soft}
    M15 - \textbf{Closed book decoder} & 3.35 & \textbf{4.95} & 4.80 & 3.67 \\ \hline %  \citep{jiang-bansal-2018-closed}
    M17 - \textbf{T5} & 4.00 & 4.93 & \textbf{4.93} & 4.23   \\ \hline %   \citep{2019t5}
    M20 - \textbf{GPT-2} (zero shot)\footnotemark  & 3.63 & 3.40 & 3.97 & 3.30   \\ \hline %  \citep{ziegler2019fine}
    M22 - \textbf{BART} & \textbf{4.18} & 4.94 & \textbf{4.90} & \textbf{4.25}  \\ \hline %  \citep{lewis2019bart}
    M23 - \textbf{Pegasus} (C4) & \textbf{4.16} & 4.91 & \textbf{4.88} & \textbf{4.26} \\ \hline %  \citep{zhang2019pegasus}
  M23 - \textbf{Pegasus} (dynamic mix) & 4.09 & 4.85 & 4.79 & \textbf{4.27} \\ \hline %  \citep{zhang2019pegasus}
  \end{tabular}
  \end{adjustbox}
  \caption{Human ratings of summaries along four evaluation dimensions, averaged over three expert annotators, broken down by extractive and abstractive models.
  The M* codes follow the notation described in Section~\ref{sec:metrics-models-models}. 
  The three highest-rated models in each column are in bold.}
  \label{models_human_judgments}
\end{table}
}

\newcommand{\ModelBenchmarkTableA}{
\begin{table*}[h!]
\begin{subtable}{\textwidth}
\centering
\small

\begin{adjustbox}{max width=\linewidth}
\begin{tabular}{l|c|c|c|c|c|c|c|c|c}
\textbf{Method} & ROUGE-1/2/3/4/L/su*/w & ROUGE-WE-(1/2/3) & $S^3$ (pyr/resp) & BertScore & MoverScore & SummaQA & SMS & BLANC & SUPERT \\ 
\hline
\multicolumn{10}{c}{\textit{Extractive Models}} \\ \hline
M0 - \textbf{LEAD-3} & 0.3994 / 0.1746 / 0.0990 / 0.0647 / 0.3606 / 0.1377 / 0.2072 & 0.4049 / 0.2260 / 0.2172 & \textbf{0.5395} / \textbf{0.6328} & 0.3742 & 0.1679 & \textbf{0.1652} & 0.1050 & 0.0480 &	\textbf{0.7259} \\ \hline
M1 - \textbf{NEUSUM} & 0.4130 / 0.1893 / 0.1109 / 0.0742 / 0.3768 / 0.1495 / 0.2156 & 0.4186 / 0.2402 / 0.2310 & \textbf{0.5562} / \textbf{0.6509} & 0.3955 & 0.1839 & \textbf{0.1700} & \textbf{0.1062} & \textbf{0.1087} & 0.7010 \\ \hline % \citep{zhou-etal-2018-neural} 
M2 - \textbf{BanditSum} & 0.4137 / 0.1868 / 0.1086 / 0.0721 / 0.3759 / 0.1513 / 0.2139 & 0.4195 / 0.2385 / 0.2300 & \textbf{0.5339} / 0.6306 & 0.3938 & 0.1815 & 0.1324 & \textbf{0.1058} & 0.0909 & \textbf{0.7018} \\ \hline % \citep{dong-etal-2018-banditsum} 
M3 - \textbf{LATENT} & 0.4136 / 0.1867 / 0.1085 / 0.0721 / 0.3757 / 0.1512 / 0.2138 & 0.4194 / 0.2384 / 0.2299 & 0.5337 / 0.6305 & 0.3936 & 0.1814 & \textbf{0.1645} & 0.1058 & 0.0910 & \textbf{0.7020} \\ \hline % \citep{zhang-etal-2018-neural-latent}
M4 - \textbf{REFRESH}  & 0.3972 / 0.1807 / 0.1042 / 0.0690 / 0.3621 / 0.1340 / 0.2129 & 0.4023 / 0.2318 / 0.2238 & \textbf{0.6395} / \textbf{0.7124} & 0.3903 & 0.1720 & \textbf{0.1944} & \textbf{0.1088} & \textbf{0.1406} & \textbf{0.7526} \\ \hline % \citep{narayan-etal-2018-ranking}
M5 - \textbf{RNES} & 0.4088 / 0.1878 / 0.1102 / 0.0736 / 0.3719 / 0.1446/ \textbf{0.2163} & 0.4153 / 0.2395 / 0.2317 & \textbf{0.6082} / \textbf{0.6894} & \textbf{0.3997} & 0.1802 & \textbf{0.1794} & \textbf{0.1107} & \textbf{0.1232}	 & \textbf{0.7434} \\ \hline % \citep{wu2018learning}
M6 - \textbf{JECS} &     0.4144 / 0.1846 / 0.1063 / 0.0699 / 0.3760 / 0.1485 / 0.2135 & 0.4200 / 0.2371 / 0.2283 & 0.5337 / 0.6284 & 0.3925 & 0.1805 & 0.1644 & 0.1048 & \textbf{0.1044} &	0.6946 \\ \hline % \citep{xu-durrett-2019-neural}
M7 - \textbf{STRASS}  &     0.3377 / 0.1237 / 0.0650 / 0.0416 / 0.2790 / 0.1052 / 0.1559 & 0.3477 / 0.1757 / 0.1656 & 0.3632 / 0.4939 & 0.3090 & 0.1079 & 0.1367 & 0.1023 & \textbf{0.1042} & 0.6566 \\ \hline %  \citep{bouscarrat-etal-2019-strass}
  \multicolumn{10}{c}{\textit{Abstractive Models}} \\ \hline
M8 - \textbf{Pointer Generator }  &  0.3921 / 0.1723 / 0.1003 / 0.0674 / 0.3599 / 0.1435 / 0.1999 & 0.3990 / 0.2226 / 0.2128 & 0.4328 / 0.5561 & 0.3763 & 0.1643 & 0.1398 & 0.0974 & 0.0704	& 0.6501 \\ \hline % \citep{see-etal-2017-get} 
M9 - \textbf{Fast-abs-rl} &       0.4057 / 0.1774 / 0.0975 / 0.0616 / 0.3806 / 0.1439 / 0.2112 & 0.4123 / 0.2302 / 0.2184 & 0.4818 / 0.5865 & 0.3918 & 0.1748 & 0.1431 & 0.0847 & 0.0855 & 0.6125 \\ \hline % \citep{chen-bansal-2018-fast} 
M10 - \textbf{Bottom-Up} &    0.4124 / 0.1870 / 0.1064 / 0.0695 / \textbf{0.3815} / 0.1543 / 0.2084 & 0.4192 / 0.2400 / 0.2313 & 0.4450 / 0.5655 & 0.3964 & 0.1830 & 0.1408 & 0.0925 & 0.0570 & 0.6092 \\ \hline % \citep{gehrmann-etal-2018-bottom} 
M11 - \textbf{Improve-abs} &     0.3985 / 0.1720 / 0.0927 / 0.0567 / 0.3730 / 0.1431 / 0.2073 & 0.4045 / 0.2300 / 0.2228 & 0.4899 / 0.5897 & 0.3826 & 0.1652 & 0.1341 & 0.0816 & 0.0777 & 0.5972 \\ \hline %  \citep{kryscinski-etal-2018-improving}
M12 - \textbf{Unified-ext-abs} &  0.4038 / 0.1790 / 0.1039 / 0.0695 / 0.3675 / 0.1484 / 0.2074 & 0.4097 / 0.2299 / 0.2204 & 0.4936 / 0.5995 & 0.3832 & 0.1739 & 0.1530 & 0.1038 & 0.0962 & 0.6826 \\ \hline %  \citep{hsu-etal-2018-unified}
M13 - \textbf{ROUGESal} &   0.4016 / 0.1797 / 0.1053 / 0.0709 / 0.3679 / 0.1497 / 0.2058 & 0.4078 / 0.2294 / 0.2190 & 0.4643 / 0.5799 & 0.3837 & 0.1722 & 0.1475 & 0.1009 & 0.0882 & 0.6570 \\ \hline %   \citep{pasunuru-bansal-2018-multi}
M14 - \textbf{Multi-task (Ent + QG )} &      0.3952 / 0.1758 / 0.1037 / 0.0705 / 0.3625 / 0.1476 / 0.2007 & 0.4015 / 0.2253 / 0.2149 & 0.4246 / 0.5513 & 0.3759 & 0.1670 & 0.1360 & 0.0982 & 0.0648 & 0.6380 \\ \hline %  \citep{guo-etal-2018-soft}
M15 - \textbf{Closed book decoder} & 0.3976 / 0.1760 / 0.1031 / 0.0696 / 0.3636 / 0.1472 / 0.2033 & 0.4039 / 0.2263 / 0.2160 & 0.4591 / 0.5757 & 0.3783 & 0.1699 & 0.1456 & 0.1009 & 0.0896 & 0.6612 \\ \hline %  \citep{jiang-bansal-2018-closed}
M16 - \textbf{SENECA }  &     0.4151 / 0.1836 / 0.1052 / 0.0681 / 0.3806 / 0.1520 / 0.2112 & 0.4211 / 0.2369 / 0.2282 & 0.4735 / 0.5836 & 0.3907 & 0.1811 & 0.1404 & 0.1005 & 0.0692 & 0.6519 \\ \hline %  \citep{sharma-etal-2019-entity}
M17 -  \textbf{T5} & \textbf{0.4479} / \textbf{0.2205} / \textbf{0.1336} / \textbf{0.0920} / \textbf{0.4172} / \textbf{0.1879} / \textbf{0.2291} & \textbf{0.4543} / \textbf{0.2723} / \textbf{0.2631} & 0.5168 / 0.6294 & \textbf{0.4450} & \textbf{0.2376} & 0.1437 & 0.1046 & 0.0773 & 0.6094 \\ \hline %   \citep{2019t5}
M18 - \textbf{NeuralTD} &        0.4004 / 0.1762 / 0.1000 / 0.0650 / 0.3723 / 0.1452 / 0.2085 & 0.4063 / 0.2277 / 0.2187 & 0.4946 / 0.5975 & 0.3949 & 0.1697 & 0.1440 & 0.0916 & 0.0859 & 0.6290 \\ \hline %  \citep{bohm-etal-2019-better}
M19 - \textbf{BertSum-abs} &     \textbf{0.4163} / \textbf{0.1944} / \textbf{0.1156} / \textbf{0.0785} / 0.3554 / \textbf{0.1625} / 0.1979 & \textbf{0.4230} / \textbf{0.2454} / \textbf{0.2351} & 0.4664 / 0.5855 & 0.3855 & \textbf{0.1894} & 0.1385 & \textbf{0.1071} & 0.0815 & 0.6116 \\ \hline %  \citep{liu-lapata-2019-text}
M20 - \textbf{GPT-2} (supervised) &     0.3981 / 0.1758 / 0.0993 / 0.0649 / 0.3674 / 0.1470 / 0.2006 & 0.4048 / 0.2268 / 0.2170 & 0.4069 / 0.5373 & 0.3915 & 0.1750 & 0.1299 & 0.0930 & 0.0705 & 0.6053 \\ \hline % \citep{ziegler2019fine} 
M21 - \textbf{UniLM} &  \textbf{0.4306} / \textbf{0.2044} / \textbf{0.1218} / \textbf{0.0824} / \textbf{0.4013} / \textbf{0.1714} / \textbf{0.2228} & \textbf{0.4369} / \textbf{0.2567} / \textbf{0.2483} & 0.5143 / 0.6210 & \textbf{0.4122} & \textbf{0.2112} & 0.1455 & 0.0957 & 0.0841 & 0.6100 \\ \hline %  \citep{dong2019unified}
M22 - \textbf{BART} &    \textbf{0.4416} / \textbf{0.2128} / \textbf{0.1285} / \textbf{0.0880} / \textbf{0.4100} / \textbf{0.1818} / \textbf{0.2266} & \textbf{0.4472} / \textbf{0.2646} / \textbf{0.2556} & 0.5116 / 0.6215 & \textbf{0.4264} & \textbf{0.2259} & 0.1457 & 0.1037 & 0.0822 & 0.6184 \\ \hline %  \citep{lewis2019bart}
M23 - \textbf{Pegasus} (dynamic mix) &   \textbf{0.4407} / \textbf{0.2155} / \textbf{0.1307} / \textbf{0.0901} / \textbf{0.4101} / \textbf{0.1825} / \textbf{0.2260} & \textbf{0.4471} / \textbf{0.2668} / \textbf{0.2575} & 0.5099 / \textbf{0.6233} & \textbf{0.4369} & \textbf{0.2283} & 0.1422 & 0.1040 & 0.0797 & 0.6046 \\ \hline %  \citep{zhang2019pegasus}
M23 - \textbf{Pegasus} (huge news) &   \textbf{0.4408} / \textbf{0.2147} / \textbf{0.1295} / \textbf{0.0889} / \textbf{0.4103} / \textbf{0.1821} / \textbf{0.2273} & \textbf{0.4473} / \textbf{0.2663} / \textbf{0.2568} & 0.5295 / \textbf{0.6372} & \textbf{0.4377} & \textbf{0.2286} & 0.1497 & 0.1049 & 0.0845 & 0.6148 \\ \hline %  \citep{zhang2019pegasus}
\end{tabular}
\end{adjustbox}
\caption{Model scores from summarization-specific evaluation metrics.}
\end{subtable}

\vspace{6pt}

\begin{subtable}{\textwidth}
\centering
\tiny
\begin{tabular}{l|c|c|c|c|c|c|c}
\textbf{Method} &  BLEU & CHRF & CIDEr & METEOR & Length & Stats (cov/comp/den) & Repeated (1/2/3) \\ \hline
\multicolumn{8}{c}{\textit{Extractive Models}} \\ \hline
M0 - \textbf{LEAD-3} &  11.4270 & 0.3892 & 0.2125 & \textbf{0.2141} & 87.4475 & 0.9825 / 9.6262 / 57.8001 & 0.2086 / 0.0310 / 0.0310 \\ \hline
M1 - \textbf{NEUSUM} &    12.7784 & 0.3946 & 0.2832 & \textbf{0.2183} & 84.4075 & 0.9819 / 9.8047 / 32.8574 & 0.2325 / 0.0531 / 0.0531 \\ \hline % \citep{zhou-etal-2018-neural}
M2 - \textbf{BanditSum} &      12.9761 & 0.3897 & 0.3305 & 0.2124 & 78.5279 & 0.9836 / 10.2810 / 40.4265 & 0.2384 / 0.0573 / 0.0573 \\ \hline % \citep{dong-etal-2018-banditsum}
M3 - \textbf{LATENT} &   12.9725 & 0.3897 & 0.3305 & 0.2123 & 78.5279 & 0.9834 / 10.2809 / 40.4095 & 0.2384 / 0.0573 / 0.0573 \\ \hline % \citep{zhang-etal-2018-neural-latent}
M4 - \textbf{REFRESH}  &     10.6568 & \textbf{0.4526} & 0.0677 & \textbf{0.2395} & 114.5684 & 0.9850 / 7.1059 / 53.1928 & 0.2127 / 0.0289 / 0.0289 \\ \hline % \citep{narayan-etal-2018-ranking}
M5 - \textbf{RNES} &     11.2203 & \textbf{0.4062} & 0.1559 & \textbf{0.2300} & 99.9199 & 0.9938 / 7.9032 / 67.7089 & 0.2451 / 0.0540 / 0.0540 \\ \hline % \citep{wu2018learning}
M6 - \textbf{JECS} &     12.5659 & \textbf{0.4310} & 0.3090 & 0.2122 & 79.7797 & 0.9874 / 10.1111 / 26.6943 & 0.2041 / 0.0327 / 0.0327 \\ \hline % \citep{xu-durrett-2019-neural}
M7 - \textbf{STRASS}  &     7.8330 & 0.3330 & 0.2945 & 0.1607 & 76.4859 & 0.9969 / 12.7835 / 59.9498 & 0.1864 / 0.0343 / 0.0343 \\ \hline % \citep{bouscarrat-etal-2019-strass}
   \multicolumn{8}{c}{\textit{Abstractive Models}} \\ \hline
M8 - \textbf{Pointer Generator }  &  13.8247 & 0.3567 & 0.5065 & 0.1860 & 63.5211 & 0.9957 / 13.1940 / 26.0880 & 0.2015 / 0.0375 / 0.0375 \\ \hline % \citep{see-etal-2017-get}
M9 - \textbf{Fast-abs-rl} &       12.9812 & 0.3778 & 0.4329 & 0.2014 & 70.8600 & 0.9860 / 11.0141 / \textbf{9.9859} & 0.2157 / 0.0370 / 0.0370 \\ \hline % \citep{chen-bansal-2018-fast}
M10 - \textbf{Bottom-Up} &    \textbf{15.1293} & 0.3523 & \textbf{0.6176} & 0.1887 & \textbf{56.5715} & 0.9811 / \textbf{14.7771} / 12.6181 & \textbf{0.1856} / \textbf{0.0211} / \textbf{0.0211} \\ \hline %  \citep{gehrmann-etal-2018-bottom}
M11 - \textbf{Improve-abs} &     11.9816 & 0.3715 & 0.3356 & 0.2005 & 75.9512 & \textbf{0.9674} / 10.6043 / \textbf{8.9755} & 0.2499 / 0.0542 / 0.0542 \\ \hline %  \citep{kryscinski-etal-2018-improving}
M12 - \textbf{Unified-ext-abs} &  12.8457 & 0.3786 & 0.3851 & 0.2017 & 74.4663 & 0.9868 / 10.7510 / 33.1106 & 0.2177 / 0.0493 / 0.0493 \\ \hline % \citep{hsu-etal-2018-unified}
M13 - \textbf{ROUGESal} &   13.8882 & 0.3668 & 0.4746 & 0.1936 & 66.5575 & 0.9853 / 13.0369 / 25.2893 & 0.2102 / 0.0458 / 0.0458 \\ \hline % \citep{pasunuru-bansal-2018-multi}
M14 - \textbf{Multi-task (Ent + QG )} &      14.5276 & 0.3539 & 0.5749 & 0.1831 & \textbf{60.0294} & 0.9853 / \textbf{14.1828} / 22.2296 & 0.1985 / 0.0411 / 0.0411 \\ \hline % \citep{guo-etal-2018-soft} 
M15 - \textbf{Closed book decoder} & 13.4158 & 0.3675 & 0.4648 & 0.1925 & 68.2858 & 0.9866 / 12.0588 / 27.3686 & 0.2074 / 0.0444 / 0.0444 \\ \hline % \citep{jiang-bansal-2018-closed} 
M16 - \textbf{SENECA } &     13.7676 & 0.3660 & 0.5233 & 0.1966 & 64.9710 & 0.9880 / 12.3610 / 16.7640 & 0.2146 / 0.0303 / 0.0303 \\ \hline % \citep{sharma-etal-2019-entity} 
M17 -  \textbf{T5} & \textbf{19.3891} & 0.3833 & \textbf{0.7763} & 0.2140 & \textbf{59.5288} & \textbf{0.9775} / \textbf{14.2002} / 12.9565 & \textbf{0.1810} / \textbf{0.0209} / \textbf{0.0209} \\ \hline % \citep{2019t5} 
M18 - \textbf{NeuralTD} &        12.9241 & 0.3783 & 0.3543 & 0.2038 & 74.4033 & 0.9830 / 10.7768 / \textbf{12.4443} & 0.2645 / 0.0901 / 0.0901 \\ \hline %  \citep{bohm-etal-2019-better}
M19 - \textbf{BertSum-abs} &     14.9525 & 0.3649 & \textbf{0.6240} & 0.1876 & \textbf{60.8893} & \textbf{0.9517} / \textbf{13.9197} / \textbf{12.3254} & \textbf{0.1697} / \textbf{0.0156} / \textbf{0.0156} \\ \hline % \citep{liu-lapata-2019-text} 
M20 - \textbf{GPT-2} (supervised) &     13.9364 & 0.3678 & 0.5787 & 0.1759 & \textbf{51.8352} & 0.9791 / \textbf{15.9839} / 15.4999 & 0.1875 / 0.0362 / 0.0362 \\ \hline % \citep{ziegler2019fine}
M21 - \textbf{UniLM} &  \textbf{15.5736} & \textbf{0.4230} & 0.5294 & 0.2084 & 67.1960 & \textbf{0.9685} / 11.5672 / \textbf{11.7908} & \textbf{0.1722} / \textbf{0.0180} / \textbf{0.0180} \\ \hline %  \citep{dong2019unified}
M22 - \textbf{BART} &    \textbf{17.1005} & \textbf{0.4271} & \textbf{0.7573} & 0.2105 & 62.2989 & \textbf{0.9771} / 12.8811 / 15.2999 & \textbf{0.1627} / \textbf{0.0127} / \textbf{0.0127} \\ \hline %  \citep{lewis2019bart}
M23 - \textbf{Pegasus} (dynamic mix) &   \textbf{18.6517} & 0.4261 & \textbf{0.7280} & \textbf{0.2131} & 64.1348 & 0.9438 / 13.7208 / 11.6003 & 0.1855 / 0.0355 / 0.0081 \\ \hline % \citep{zhang2019pegasus}
M23 - \textbf{Pegasus} (huge news) &   \textbf{17.8102} & 0.3912 & \textbf{0.6595} & \textbf{0.2189} & 66.7559 & 0.9814/12.9473/14.9850 & 0.1883/0.0251/0.0251 \\ \hline % \citep{zhang2019pegasus}
\end{tabular}
\caption{Model scores from other text generation evaluation metrics.}
\end{subtable}
\caption{Model scores from automatic evaluation metrics available in the evaluation toolkit. The five highest scores for each metric (and lowest for Length and Repeated-1/2/3) are bolded.}
\label{model_scores_1}
\end{table*}
}

%% file: helpers-new-tabs-figs.tex
\newcommand{\HistogramAgreementCombined}{
\begin{figure*}[t!]
    \centering
     \includegraphics[width=\linewidth]{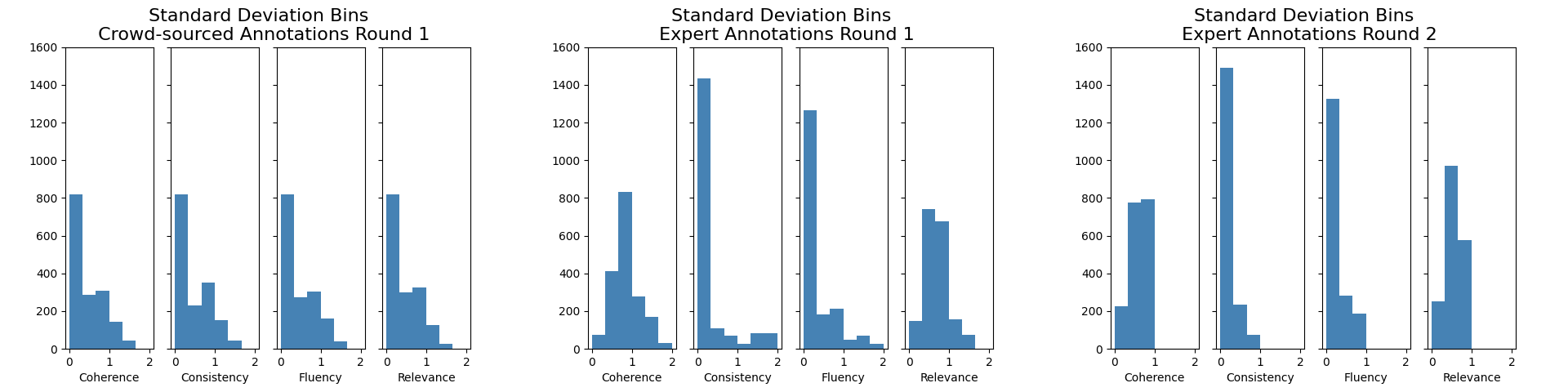}
    \caption{Histogram of standard deviations of inter-annotator scores between: crowd-sourced annotations, first round expert annotations, second round expert annotations, respectively.}
    \label{fig:agreement_histogram_all}
\end{figure*} 
}

\newcommand{\MetricPairwiseCorrelationsKendall}{
\begin{figure*}[ht]
    \centering
    \includegraphics[width=0.8\linewidth]{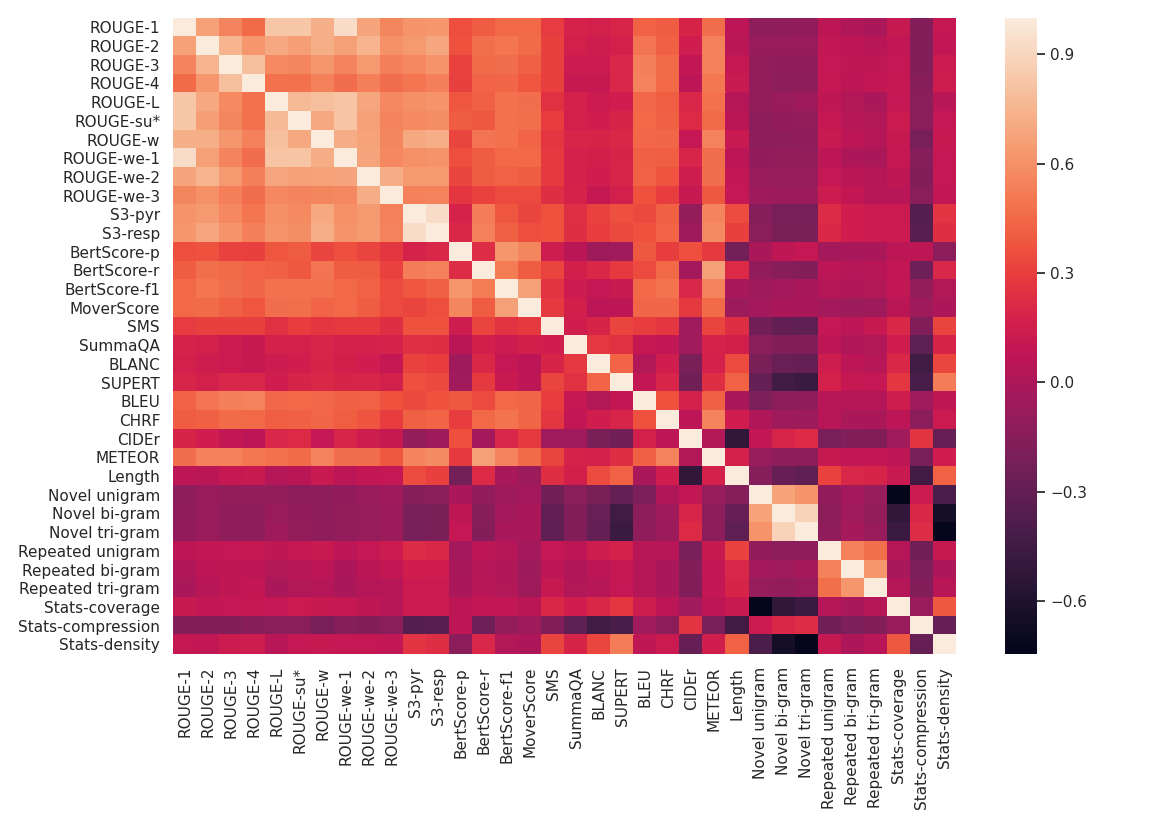}
    \caption{Pairwise Kendall's Tau correlations for all automatic evaluation metrics.
    }
    \label{fig:pairwise_correlations_kendall}
\end{figure*} 
}

\newcommand{\DataCollectionInterface}{
\begin{figure*}[hb]
    \centering
    \includegraphics[width=0.75\linewidth]{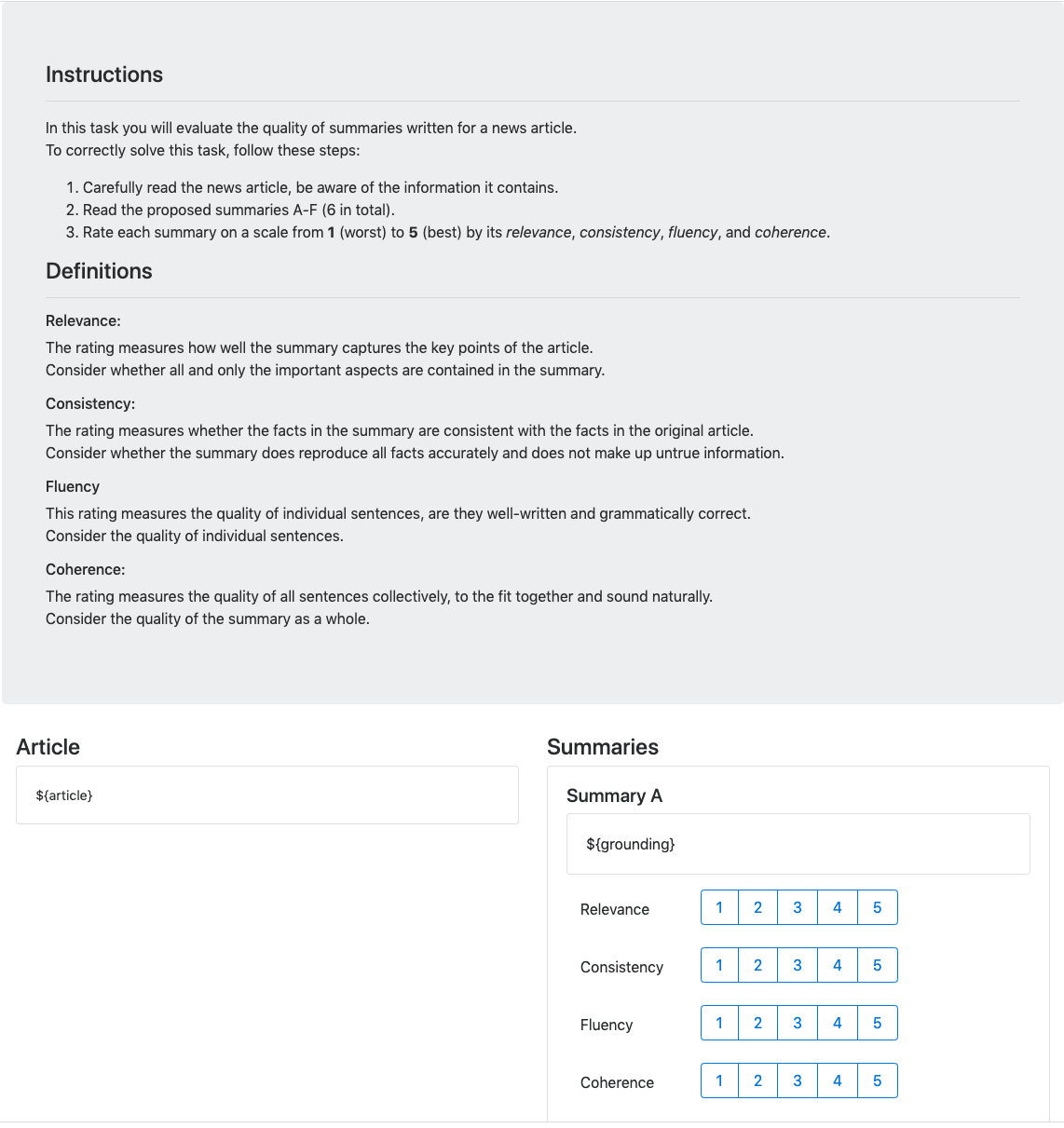}
    \caption{Example of the data collection interface used by crowd-source and expert annotators.}
    \label{fig:data_collection_ui}
\end{figure*} 
}

\newcommand{\MetricKendallCorrelationTableSystemLevel}{
\begin{table}[t!]
\centering
\small
\begin{adjustbox}{max width=\linewidth}
\begin{tabular}{l|c|c|c|c}
\textbf{Metric} & Coherence & Consistency & Fluency & Relevance \\ \hline
ROUGE-1   &  0.2500  &  0.5294  & \textbf{ 0.5240 } &  0.4118  \\ \hline
ROUGE-2   &  0.1618  &  0.5882  &  0.4797  &  0.2941  \\ \hline
ROUGE-3    &  0.2206  & \textbf{ 0.7059 } & \textbf{ 0.5092 } &  0.3529  \\ \hline
ROUGE-4   & \textbf{ 0.3088 } &  0.5882  & \textbf{ 0.5535 } &  0.4118  \\ \hline
ROUGE-L    &  0.0735  &  0.1471  &  0.2583  &  0.2353  \\ \hline
ROUGE-su*   &  0.1912  &  0.2941  &  0.4354  &  0.3235  \\ \hline
ROUGE-w    &  0.0000  &  0.3971  &  0.3764  &  0.1618  \\ \hline
ROUGE-we-1   &  \textbf{0.2647}  &  0.4559  & \textbf{ 0.5092 } & \textbf{\textbf{ 0.4265  }}\\ \hline
ROUGE-we-2   &  -0.0147  &  0.5000  &  0.3026  &  0.1176  \\ \hline
ROUGE-we-3    &  0.0294  &  0.3676  &  0.3026  &  0.1912  \\ \hline
$S^3$-pyr   &  -0.0294  &  0.5147  &  0.3173  &  0.1324  \\ \hline
$S^3$-resp    &  -0.0147  &  0.5000  &  0.3321  &  0.1471  \\ \hline
BertScore-p   &  0.0588  &  -0.1912  &  0.0074  &  0.1618  \\ \hline
BertScore-r   &  0.1471  & \textbf{ 0.6618 } &  0.4945  &  0.3088  \\ \hline
BertScore-f    &  0.2059  &  0.0441  &  0.2435  & \textbf{\textbf{ 0.4265  }}\\ \hline
MoverScore   &  0.1912  &  -0.0294  &  0.2583  &  0.2941  \\ \hline
SMS    &  0.1618  &  0.5588  &  0.3616  &  0.2353  \\ \hline
SummaQA\^   &  0.1176  &  \textbf{0.6029}  &  0.4059  &  0.2206  \\ \hline
BLANC\^  &  0.0735  &  0.5588  &  0.3616  &  0.2647  \\ \hline
SUPERT\^  &  0.1029  &  0.5882  &  0.4207  &  0.2353  \\ \hline
BLEU    &  0.1176  &  0.0735  &  0.3321  &  0.2206  \\ \hline
CHRF    & \textbf{ 0.3971 } &  0.5294  &  0.4649  & \textbf{ 0.5882  }\\ \hline
CIDEr    &  0.1176  &  -0.1912  &  -0.0221  &  0.1912  \\ \hline
METEOR    &  0.2353  & \textbf{ 0.6324 } & \textbf{ 0.6126 } & \textbf{\textbf{ 0.4265  }}\\ \hline
Length\^    &  -0.0294  &  0.4265  &  0.2583  &  0.1618  \\ \hline
Novel unigram\^   &  0.1471  &  -0.2206  &  -0.1402  &  0.1029  \\ \hline
Novel bi-gram\^   &  0.0294  &  -0.5441  &  -0.3469  &  -0.1029  \\ \hline
Novel tri-gram\^    &  0.0294  &  -0.5735  &  -0.3469  &  -0.1324  \\ \hline
Repeated unigram\^   & \textbf{\textbf{ -0.3824 }} &  0.1029  &  -0.0664  &  -0.3676  \\ \hline
Repeated bi-gram\^   & \textbf{\textbf{ -0.3824 }} &  -0.0147  &  -0.2435  & \textbf{ -0.4559  }\\ \hline
Repeated tri-gram\^    &  -0.2206  &  0.1471  &  -0.0221  &  -0.2647  \\ \hline
Stats-coverage\^   &  -0.1324  &  0.3529  &  0.1550  &  -0.0294  \\ \hline
Stats-compression\^   &  0.1176  &  -0.4265  &  -0.2288  &  -0.0147  \\ \hline
Stats-density\^   &  0.1618  & \textbf{ 0.6471 } &  0.3911  &  0.2941  \\ \hline
  \end{tabular}
  \end{adjustbox}
  \caption{Kendall's tau correlation coefficients of expert annotations computed on a system-level along four quality dimensions with automatic metrics using 11 reference summaries per example.
  \^{}~denotes metrics which use the source document.
  The five most-correlated metrics in each column are bolded.}
   \label{correlations_11_system}
\end{table}
}

%% file: 0-abstract.tex
    \begin{abstract}
The scarcity of comprehensive up-to-date studies on evaluation metrics for text summarization and the lack of consensus regarding evaluation protocols continue to inhibit progress.
We address the existing shortcomings of summarization evaluation methods along five dimensions:
1)~we re-evaluate 14 automatic evaluation metrics in a comprehensive and consistent fashion using neural summarization model outputs along with expert and crowd-sourced human annotations,
2)~we consistently benchmark 23 recent summarization models using the aforementioned automatic evaluation metrics,
3)~we assemble the largest collection of summaries generated by models trained on the CNN/DailyMail news dataset and share it in a unified format,
4)~we implement and share a toolkit that provides an extensible and unified API for evaluating summarization models across a broad range of automatic metrics,  
5)~we assemble and share the largest and most diverse, in terms of model types, collection of human judgments of model-generated summaries on the CNN/Daily Mail dataset annotated by both expert judges and crowd-source workers.
% ALEX: modified the above because this paper (https://www.aclweb.org/anthology/P18-1060.pdf) releases a large scale judgment dataset (also available in sacrerouge) but only available for 4 models. Also just crowdsourced and not expert so could emphasize that.  
%
We hope that this work will help promote a more complete evaluation protocol for text summarization as well as advance research in developing evaluation metrics that better correlate with human judgments.
\end{abstract}

%Automatic text summarization has seen a recent increase in popularity and improved results, especially with the application of pretrained language models, for both extractive and abstractive approaches. 
% 
% DONE - BRYAN: Best not start out the first sentence everyone will read by saying y'all are doing it b/c its a popular thing to do -- lead with something more related to the potential impact that summarization (or even better, this work) can have rather than its popularity.
%Automatic text summarization has seen a recent increase in popularity and improved results, especially with the application of pretrained language models, for both extractive and abstractive approaches. 
%DONE -  BRYAN: My sense here is that this isn't really a benchmark for summariation models, though perhaps it is for evaluation metrics themselves?
%However, the lack of a proper benchmark and human-correlated evaluation metrics continues to inhibit progress in the field.
%
%We present SUMM-EVAL to address existing shortcomings along three dimensions:
%
%1) a collection of  on the CNN-Daily Mail news summarization dataset from 23 recent papers in a unified format
%
%2) the release of a toolkit for summarization evaluation which includes 12 automatic metrics and
%
%3) the collection and analysis of human judgments of these collected models along four dimensions, namely coherence, consistency, fluency and relevance.
%
%We hope that this work will help promote a more complete comparison of English news summarization models as well as advance research in developing human-correlated evaluation metrics. 

%% file: 1-introduction.tex
\section{Introduction}\label{sec:introduction}
Text summarization aims to compress long document(s) into a short, fluent, and human-readable form that preserves the most salient information from the source document.

The field has benefited from advances in neural network architectures~\cite{sutskever2014sequence, bahdanau2014neural, vinyals2015pointer, vaswani2017attention} as well as the availability of large-scale datasets~\cite{sandhaus2008new, hermann2015teaching, grusky-etal-2018-newsroom, narayan-etal-2018-ranking}. 
Recent advances in pretrained language models, such as BERT~\cite{devlin-etal-2019-bert}, have motivated a corresponding shift to pretraining methods in summarization~\citep{liu-lapata-2019-text, zhang-etal-2019-hibert, dong2019unified, ziegler2019fine, 2019t5, lewis2019bart}.

A standard dataset for training summarization models is the CNN/DailyMail corpus~\cite{hermann2015teaching}, originally a question answering task, which was repurposed for summarization by~\citet{nallapati2016abstractive}.
The dataset consists of news articles and associated human-created bullet-point summaries.
The ROUGE~\citep{lin-2004-rouge} metric, which measures lexical overlap between generated and target summaries, is then typically used together with crowd-sourced human annotations for model evaluation.
While the current setup has become standardized, we believe several factors prevent a more complete comparison of models, thus negatively impacting the progress of the field.

\par
As noted by~\citet{hardy-etal-2019-highres}, recent papers vastly differ in their evaluation protocol.
Existing work often limits model comparisons to only a few baselines and offers human evaluations which are largely inconsistent with prior work.
Additionally, despite problems associated with ROUGE when used outside of its original setting~\citep{liu-liu-2008-correlation, cohan-goharian-2016-revisiting} as well as the introduction of many variations on ROUGE~\citep{zhou-etal-2006-paraeval,ng-abrecht-2015-better, ganesan2015rouge, shafieibavani-etal-2018-graph} and other text generation metrics~\citep{peyrard-2019-studying, zhao-etal-2019-moverscore, bert-score, scialom-etal-2019-answers, clark-etal-2019-sentence}, ROUGE has remained the default automatic evaluation metric.
We believe that the shortcomings of the current evaluation protocol are partially caused by the lack of easy-to-use resources for evaluation, both in the form of simplified evaluation toolkits and large collections of model outputs.

\par
In parallel, there is an issue with how evaluation metrics are evaluated themselves.
Many of the currently used metrics were developed and assessed using the Document Understanding Conference (DUC) and Text Analysis Conference (TAC) shared-tasks datasets~\cite{dang2008overview, dang2009overview}.
However, it has recently been shown that the mentioned datasets contain human judgments for model outputs scoring on a lower scale compared to current summarization systems putting into question the true performance of those metrics in the new setting~\citep{peyrard-2019-studying}. 

\par
We address these gaps in complementary ways:
1)~We re-evaluate 14 automatic evaluation metrics in a comprehensive and consistent fashion using outputs from recent neural summarization models along with expert and crowd-sourced human annotations,
2)~We consistently benchmark 23 recent summarization models using the aforementioned automatic evaluation metrics,
3) We release aligned summarization model outputs from 23 papers (44 model outputs) published between 2017 and 2019 trained on the CNN/DailyMail dataset to allow for large-scale comparisons of recent summarization models,
4) We release a toolkit of 14 evaluation metrics with an extensible and unified API to promote the reporting of additional metrics in papers,
5) We collect and release expert, as well as crowd-sourced, human judgments for 16 model outputs on 100 articles over 4 dimensions to further research into human-correlated evaluation metrics. 
Code and data associated with this work is available at~\url{https://github.com/Yale-LILY/SummEval}.
%[ANON]~\url{https://github.com/Yale-LILY/SummEval}. 

% (Alex-DONE) BRYAN: metric scores sound strange
%We report scores for all evaluation metrics over all models as well as study correlations between metrics and human judgments.
%
%We believe that a simplified toolkit for all metrics would promote the use of metrics other than ROUGE.
%We believe this gap is in part due to the lack of a unified output format and the availability of a large repository of model outputs.

%% file: 2-related-work.tex
\section{Related Work}\label{sec:related-work}
\AnnotationExamplesTable
Previous work examining the research setup of text summarization can be broadly categorized into three groups, based on the subject of analysis: evaluation metrics, datasets, and models.

\par
Dealing with evaluation methods, \citet{lin2004looking} examined the effectiveness of the ROUGE metric in various DUC tasks.
The authors concluded that evaluating against multiple references results in higher correlation scores with human judgments, however, a single-reference setting is sufficient for the metric to be effective.
\citet{owczarzak2012} studied the effects of inconsistencies in human annotations on the rankings of evaluated summarization systems.
Results showed that system-level rankings were robust against annotation inconsistencies, however, summary-level rankings were not stable in such settings and largely benefit from improving annotator consistency.
\citet{rankel-etal-2013-decade} analyzed the performance of different variants of the ROUGE metric using TAC datasets.
The authors found that higher-order and less commonly reported ROUGE settings showed a higher correlation with human judgments.
In a similar line of work, \citet{graham-2015-evaluating} conducted a large-scale study of the effectiveness of different ROUGE metric variants and compared it against the BLEU metric on the DUC datasets.
Its results highlighted several superior, non-standard ROUGE settings that achieved strong correlations with human judgments on model-generated summaries.
In \citep{chaganty-etal-2018-price} the authors investigated using an automatic metric to reduce the cost of human evaluation without introducing bias.
Together with the study, the authors released a set of human judgments over several model outputs, limited to a small set of model types. 
\citet{peyrard-2019-studying} showed that standard metrics are in agreement when dealing with summaries in the scoring range found in TAC summaries, but vastly differ in the higher-scoring range found in current models.
The authors reported that additional human annotations on modern model outputs are necessary to conduct a conclusive study of evaluation metrics.
\citet{hardy-etal-2019-highres} underscore the differences in approaches to human summary evaluation while proposing a highlight-based reference-less evaluation metric.
Other work has examined the problems with applying ROUGE in settings such as meeting summarization~\citep{liu-liu-2008-correlation} and summarization of scientific articles~\citep{cohan-goharian-2016-revisiting}. 
We build upon this line of research by examining the performance of several automatic evaluation methods, including ROUGE and its variants, against the performance of expert human annotators.

\par
In relation to datasets, \citet{dernoncourt-etal-2018-repository} presented a detailed taxonomy of existing summarization datasets.
The authors highlighted the differences in formats of available corpora and called for creating a unified data standard.
In a similar line of research, \citet{grusky-etal-2018-newsroom} offered a thorough analysis of existing corpora, focusing their efforts on news summarization datasets.
The authors also introduced several metrics for evaluating the extractiveness of summaries which are included in the toolkit implemented as part of this work.
\citet{kryscinski-etal-2020-evaluating} showed that news-related summarization datasets, such as CNN/DailyMail, contain strong layout biases.
The authors revealed that datasets in the current format, where each news article is associated with a single reference summary, leave the task of summarization underconstrained.
The paper also highlighted the problem of noisy, low-quality data in automatically-collected news datasets.

\par
Looking into models, \citet{zhang-etal-2018-abstractiveness} analyzed the level of abstraction of several recent abstractive summarization models.
The authors showed that word-level extractive models achieved a similar level of abstraction to fully abstractive models.
In \citep{kedzie-etal-2018-content} the authors examined the influence of various model components on the quality of content selection.
The study revealed that in the current setting the training signal is dominated by biases present in summarization datasets preventing models from learning accurate content selection.
\citet{kryscinski-etal-2020-evaluating} investigate the problem of factual correctness of text summarization models.
The authors concluded that the issue of hallucinating facts touches up to 30\% of generated summaries and list common types of errors made by generative models.
Closely related to that work, \citet{maynez2020} conducted a large-scale study of abstractive summarizers from the perspective of faithfulness.
The authors reached similar conclusions, stating that improving factual faithfulness is a critical issue in summarization.
The results also showed that currently available evaluation methods, such as ROUGE and BertScore, are not sufficient to study the problem at hand. \citet{durmus-etal-2020-feqa} and \citet{wang-etal-2020-asking} similarly examine faithfulness evaluation, both proposing question answering frameworks as a means of evaluating factual consistency. 

\par
Insights and contributions coming from our work are complementary to the conclusions of previous efforts described in this section.
To the best of our knowledge, this is the first work in neural text summarization to offer a large-scale, consistent, side-by-side re-evaluation of summarization model outputs and evaluation methods. 
We also share resources that we hope will prove useful for future work in analyzing and improving summarization models and metrics.

Shortly before publishing this manuscript a library for developing summarization metrics was released by \citet{deutsch2020sacrerouge}.
Our toolkit is complementary to their work as their toolkit includes only 3 of our 12 evaluation metrics.

% ALEX: WIP -- want to divide related works into three areas 1) analysis of summarization protocols/flaws in the setup, 2) resources/missing useful parts 3) evaluation metrics 
% In Natural Language Processing, there has been a community effort to track progress in the field through organized model comparisons such as~\href{http://nlpprogress.com/}{nlpprogress} and~\href{https://paperswithcode.com/}{paperswithcode}. 
%
%In summarization, recent work has focused on the analysis of specific model components as well as evaluation protocols.
%

% (Alex-DONE) BRYAN: You never need to say "In our work". Just start with "We ..."
%Several works have pointed to the variations in summarization evaluation as well as flaws in the current research setup. 
%
%%
%We address these issues and help standardize summarization evaluation through our collection of model outputs as well as the release of our evaluation toolkit. 

%
\par

%% file: 3-metrics-models.tex
\section{Evaluation Metrics and Summarization Models}\label{sec:metrics-models}
We briefly introduce metrics included in our evaluation toolkit as well as the summarization models for which outputs were collected at the time of releasing this manuscript. 

\subsection{Evaluation Metrics}\label{sec:metrics-models-metrics}
Our selection of evaluation methods includes several recently introduced metrics that have been applied to both text generation and summarization, standard machine translation metrics, and other miscellaneous performance statistics. 
\par
\noindent \textbf{ROUGE}~\citep{lin-2004-rouge}, (Recall-Oriented Understudy for Gisting Evaluation), measures the number of overlapping textual units (n-grams, word sequences) between the generated summary and a set of gold reference summaries.

\par
\noindent \textbf{ROUGE-WE}~\citep{ng-abrecht-2015-better} extends ROUGE by using soft lexical matching based on the cosine similarity of Word2Vec~\cite{NIPS2013_5021} embeddings.

\par
\noindent \boldmath{$S^3$}~\citep{peyrard-etal-2017-learning} is a model-based metric that uses previously proposed evaluation metrics, such as ROUGE, JS-divergence, and ROUGE-WE, as input features for predicting the evaluation score.
The model is trained on human judgment datasets from TAC conferences.

\par
\noindent \textbf{BertScore}~\citep{bert-score} computes similarity scores by aligning generated and reference summaries on a token-level.
Token alignments are computed greedily to maximize the cosine similarity between contextualized token embeddings from BERT.

\par
\noindent \textbf{MoverScore}~\citep{zhao-etal-2019-moverscore} measures the semantic distance between a summary and reference text by making use of the Word Mover's Distance~\cite{kusner2015word} operating over n-gram embeddings pooled from BERT representations.
\par
\noindent \textbf{Sentence Mover's Similarity (SMS)}~\citep{clark-etal-2019-sentence} extends Word Mover's Distance to view documents as a bag of sentence embeddings as well as a variation which represents documents as both a bag of sentences and a bag of words.

\par
\noindent \textbf{SummaQA}~\citep{scialom-etal-2019-answers} applies a BERT-based question-answering model to answer cloze-style questions using generated summaries.
Questions are generated by masking named entities in source documents associated with evaluated summaries.
The metric reports both the F1 overlap score and QA-model confidence. 

\par
\noindent \textbf{BLANC}~\citep{vasilyev-2020-blanc} is a reference-less metric which measures the performance gains of
a pre-trained language model given access to a document summary while carrying out language
understanding tasks on the source document's text. 

\par
\noindent \textbf{SUPERT}~\citep{Gao-2020-supert} is a reference-less metric, originally designed for multi-document summarization, which measures the semantic similarity of model outputs with pseudo-reference summaries created by extracting salient sentences from the source documents, using soft token alignment techniques.
% AF - I like reference-less rather than unsupervised and we used the wording reference-less in the related work section
\par 
\noindent \textbf{BLEU}~\citep{papineni-etal-2002-bleu} is a corpus-level precision-focused metric which calculates n-gram overlap between a candidate and reference utterance and includes a brevity penalty. 
It is the primary evaluation metric for machine translation.

\par
\noindent \textbf{CHRF}~\citep{popovic-2015-chrf} calculates character-based n-gram overlap between model outputs and reference documents.
%and has been used primarily in machine translation.

\par
\noindent \textbf{METEOR}~\citep{lavie-agarwal-2007-meteor} computes an alignment between candidate and reference sentences by mapping unigrams in the generated summary to 0 or 1 unigrams in the reference, based on stemming, synonyms, and paraphrastic matches.
Precision and recall are computed and reported as a harmonic mean.

\par
\noindent \textbf{CIDEr}~\citep{vedantam2015cider} computes \{1-4\}-gram co-occurrences between the candidate and reference texts, down-weighting common n-grams and calculating cosine similarity between the n-grams of the candidate and reference texts.

\par
\noindent  \textbf{Data Statistics}:~\citet{grusky-etal-2018-newsroom} define three measures of the extractiveness of a dataset.
\textit{Extractive fragment coverage} is the percentage of words in the summary that are from the source article, measuring the extent to which a summary is a derivative of a text.
\textit{Density} is defined as the average length of the extractive fragment to which each summary word belongs.
\textit{Compression ratio} is defined as the word ratio between the articles and its summaries:
In addition to these measures, we also include the percentage of n-grams in the summary not found in the input document as a {\em novelty} score and the percentage of n-grams in the summary which repeat as a score of {\em redundancy}. 
For a comprehensive explanation of each metric, please refer to the corresponding paper.

\subsection{Summarization models}\label{sec:metrics-models-models}
%\MetricPearsonCorrelationTable
\MetricKendallCorrelationTableSystemLevel
We broadly categorize the models included in this study into extractive and abstractive approaches.
For each model, we provide a model code (M*) as well as a descriptive model name which will allow for easy matching with the released data.

\subsection*{Extractive Methods}

M1 - \textbf{NEUSUM}~\citep{zhou-etal-2018-neural} jointly scores and selects sentences by first building a hierarchical representation of a document and considering the partially outputted summary at each time step. 

\par
\noindent  M2 - \textbf{BanditSum}~\citep{dong-etal-2018-banditsum} treats extractive summarization as a contextual bandit problem where the document is the context and the sequence of sentences to include in the summary is the action.

\par
\noindent M3 - \textbf{LATENT}  \citep{zhang-etal-2018-neural-latent}
propose a latent  variable  extractive  model  which  views  rele-vance labels of sentences in a document as binarylatent variables

\par
\noindent M4 - \textbf{REFRESH} \citep{narayan-etal-2018-ranking} propose using REINFORCE \cite{williams1992simple} to extract summaries, approximating the search space during training by limiting to combinations of individually high-scoring sentences. 

\par
\noindent M5 - \textbf{RNES} \citep{wu2018learning} propose a coherence model to capture cross-sentence coherence, combining output from the coherence model and ROUGE scores as a reward in a REINFORCE framework.

\par
\noindent M6 - \textbf{JECS} \citep{xu-durrett-2019-neural}
first extracts sentences from a document and then scores possible constituency-based compressed units to produce the final compressed summary. 

\par
\noindent M7 - \textbf{STRASS}~\citep{bouscarrat-etal-2019-strass}  extracts a summary by selecting the sentences with the closest embeddings to the document embedding, learning a transformation to maximize the similarity between the summary and the ground truth reference.  

\subsubsection*{Abstractive Methods}
\noindent M8 - \textbf{Pointer Generator}~\citep{see-etal-2017-get} propose a variation of encoder-decoder models, the Pointer Generator Network, where the decoder can choose to generate a word from the vocabulary or copy a word from the input.
A coverage mechanism is also proposed to prevent repeatedly attending to the same part of the source document.

\par
\noindent M9 - \textbf{Fast-abs-rl} \citep{chen-bansal-2018-fast} propose a model which first extracts salient sentences with a Pointer Network and rewrites these sentences with a  Pointer Generator Network.
In addition to maximum likelihood training, a ROUGE-L reward is used to update the extractor via REINFORCE \citep{williams1992simple}. 

\par
\noindent M10 - \textbf{Bottom-Up} \citep{gehrmann-etal-2018-bottom} introduce a bottom-up approach whereby a content selection model restricts the copy attention distribution of a pretrained Pointer Generator Network during inference.

\par
\noindent M11 - \textbf{Improve-abs} \citep{kryscinski-etal-2018-improving} extend the model of \citet{paulus2017deep} by augmenting the decoder with an external LSTM language model and add a novelty RL-based objective during training. 

\par
\noindent M12 - \textbf{Unified-ext-abs} \citep{hsu-etal-2018-unified} propose to use the probability output of an extractive model as sentence-level attention to modify word-level attention scores of an abstractive model, introducing an inconsistency loss to encourage consistency between these two levels of attention. 

\par
\noindent M13 - \textbf{ROUGESal}  \citep{pasunuru-bansal-2018-multi} propose a keyphrase-based salience reward as well as an entailment-based reward in addition to using a ROUGE-based reward in a REINFORCE setting, optimizing rewards simultaneously in alternate mini-batches. 

\par
\noindent M14 - \textbf{Multi-task (Ent + QG )} \citep{guo-etal-2018-soft} propose question generation and entailment generation as auxiliary tasks in a multi-task framework along with a corresponding multi-task architecture.

\par
\noindent   M15 - \textbf{Closed book decoder} \citep{jiang-bansal-2018-closed} build upon a Pointer Generator Network by adding copy-less and attention-less decoder during training time to force the encoder to be more selective in encoding salient content. 

\par
\noindent   M16 - \textbf{SENECA } \citep{sharma-etal-2019-entity}  propose to use entity-aware content selection module and an abstractive generation module to generate the final summary. 

 \par
\noindent    M17 -  \textbf{T5}  \citep{2019t5}  perform a systematic study of transfer learning techniques and apply their insights to a set of tasks all framed as text-input to text-output generation tasks, including summarization. 

\par
\noindent  M18 - \textbf{NeuralTD} \citep{bohm-etal-2019-better} learn a reward function from 2,500 human judgments which is used in a reinforcement learning setting. 

\par
\noindent   M19 - \textbf{BertSum-abs} \citep{liu-lapata-2019-text} introduce a novel document-level encoder on top of BERT \citep{devlin-etal-2019-bert}, over which they introduce both an extractive and an abstractive model. 

\par
\noindent M20 - \textbf{GPT-2} \citep{ziegler2019fine} build off of  GPT-2 \citep{radford2019language} and fine-tune the model by using human labels of which of four sampled summaries is the best to direct fine-tuning in a reinforcement learning framework.

\par
\noindent M21 - \textbf{UniLM} \citep{dong2019unified}  introduce a model pretrained on three language modeling tasks: unidirectional, bidirectional, and sequence-to-sequence prediction.
It is thus applicable to natural language understanding tasks and generation tasks such as abstractive summarization. 

\par
\noindent  M22 - \textbf{BART} \citep{lewis2019bart} introduce a denoising autoencoder for pretraining sequence to sequence tasks which is applicable to both natural language understanding and generation tasks.

\par
\noindent M23 - \textbf{Pegasus} \citep{zhang2019pegasus}  introduce a model pretrained with a novel objective function designed for summarization by which important sentences are removed from an input document and then generated from the remaining sentences.

% table footnote found under methodology.tex for spacing purposes

%% file: 4-methodology.tex
\section{Resources}\label{sec:methodology}
%\MetricPairwiseCorrelations
\MetricPairwiseCorrelationsKendall
We now describe the resources collected and released together with this manuscript. 
\subsection{Model Outputs}
The model output collection contains summaries associated with 23 recent papers on neural text summarization described in Section~\ref{sec:metrics-models-models}.
We obtained a total of 44 model outputs, as many papers include variations of the main model.
All models were trained on the CNN/DailyMail news corpus and the collected summaries were generated using the test split of the dataset without constraints limiting the output length.
Outputs were solicited from the authors of papers to ensure comparability between results presented in this paper with those in the original works.
They are shared publicly with the consent of the authors.

\par
Model outputs were transformed into a unified format and are shared with IDs of the original CNN/DailyMail examples so that generated summaries can be matched with corresponding source articles.
Pairing model outputs with original articles was done using a heuristic approach that relied on aligning reference summaries. 
The pairing process revealed that 38 examples in the CNN/DailyMail test split contained duplicate reference summaries preventing those examples to be correctly aligned.
However, this problem involves only 0.3\% of the available data and should not have a significant impact on downstream results.
IDs of duplicate examples are provided together with the data.

\subsection{Evaluation Toolkit}
The evaluation toolkit contains 14 automatic evaluation metrics described in Section~\ref{sec:metrics-models-metrics} consolidated into a Python package.
The package provides a high-level, easy-to-use interface unifying all of the underlying metrics.
For each metric, we implement both \texttt{evaluate\_example} and \texttt{evaluate\_batch} functions that return the metric's score on example- and corpus-levels accordingly.
Function inputs and outputs are also unified across all metrics to streamline multi-metric evaluation and result processing.
The toolkit comes with a standard configuration resembling the most popular settings for each of the metrics to enable easy, out-of-the-box use.
However, each metric can be further configured using external \texttt{gin} configuration files.
We also provide a command-line tool to evaluate a summarization model with several metrics in parallel.
%
%We believe that such a toolkit will allow for easily-comparable results; one can simply upload the output from the command-line tool along with the associated configuration file to directly compare results on the benchmark. 

\subsection{Human Annotations}\label{sec:methodology-human-annotations}
%Our overarching goal is to compare the correlation of automatic evaluation metrics with human judgments as well as compare the performance of state-of-the-art models according to these metrics and human judgments for a more complete summarization evaluation.
%\par
%We first aim to answer the call of~\citet{peyrard-2019-studying} for high quality human judgments of model summarization models by collecting human summary evaluations along the following four dimension, as in~\citet{kryscinski2019evaluating}:

The collection of human annotations contains summary evaluations of 16 recent neural summarization models solicited from crowd-sourced and expert judges. 
Annotations were collected for 100 articles randomly picked from the CNN/DailyMail test set.
To ensure high quality of annotations, each summary was scored by 5 crowd-sourced and 3 expert workers, amounting to 12800 summary-level annotations.
Model outputs were evaluated along the following four dimensions, as in~\citet{kryscinski-etal-2019-neural}:
\par
\noindent \textbf{Coherence} - the collective quality of all sentences.
We align this dimension with the DUC quality question \cite{dang2005overview} of structure and coherence whereby "the summary should be well-structured and well-organized.
The summary should not just be a heap of related information, but should build from sentence to sentence to a coherent body of information about a topic."

\par
\noindent \textbf{Consistency} - the factual alignment between the summary and the summarized source.
A factually consistent summary contains only statements that are entailed by the source document.
Annotators were also asked to penalize summaries that contained hallucinated facts.

\par
\noindent \textbf{Fluency} - the quality of individual sentences.
Drawing again from the DUC quality guidelines, sentences in the summary "should have no formatting problems, capitalization errors or obviously ungrammatical sentences (e.g., fragments, missing components) that make the text difficult to read." 

\par
\noindent \textbf{Relevance} - selection of important content from the source.
The summary should include only important information from the source document.
Annotators were instructed to penalize summaries which contained redundancies and excess information.

\par
The data collection interface provided judges with the source article and associated summaries grouped in sets of 5.
Each group of summaries contained the reference summary associated with the source article to establish a common point of reference between groups.
Summary grouping and order within groups were randomized for each annotator.
Judges were asked to rate the summaries on a Likert scale from 1 to 5 (higher better) along the four mentioned dimensions.

\par
Crowd-sourced annotators were hired through the Amazon Mechanical Turk platform.
The hiring criteria were set to a minimum of 10000 approved HITs and an approval rate of 97\% or higher.
Geographic constraints for workers were set to United States, United Kingdom, and Australia to ensure that summaries were evaluated by native English speakers.
Compensation was carefully calculated to ensure an average wage of 12 USD per hour.

\citet{gillick-liu-2010-non} showed that summary judgments obtained through non-experts may differ greatly from expert annotations and could exhibit worse inter-annotator agreement.
As a result, in addition to the hired crowd-sourced workers, we enlisted three expert annotators who have written papers on summarization either for academic conferences (2) or as part of a senior thesis (1).
The expert annotators were asked to evaluate the same set of summaries under the same instructions as the hired crowd-sourced workers.
For expert judgments, we proceeded with two rounds of annotation to correct any obvious mistakes as well as to confirm judgments and ensure a higher quality of annotations.
In the second round, annotators were asked to check all examples for which their score of a dimension differed from another annotator by more than 2 points and where the other annotators were within 1 point of each other.
In cases where a score differed by more than 2 points for which such a pattern did not exist, all annotators examined the annotation.
When re-evaluating examples, judges were allowed to see scores assigned by other expert annotators in the first round of annotations.
While such a setting could undermine the wisdom of the crowd and shift the re-assigned scores towards the average judgment from the first round, we encouraged experts to remain critical and discuss contested examples when necessary. 
For completeness, the data collection user interface and additional details regarding the data collection process are presented in the Appendix.
\footnotetext[1]{The zero-shot model was used for evaluation.}

%% file: 5-metric-evaluation.tex
\section{Metric Re-evaluation}\label{sec:metric-evaluation}
\ModelsHumanJudgementScoreTable
\subsection{Human Annotations}
Considering the concerns raised in previous work~\citep{gillick-liu-2010-non} about the quality differences between crowd-sourced and expert annotations we study this issue using the human annotations collected as part of this work.

\HistogramAgreementCombined
\par
To evaluate the inter-annotator agreement of collected crowd-sourced and expert annotations we computed the Krippendorff's alpha coefficient~\citep{krippendorff2011computing}.
We found the inter-annotator interval kappa to be below an acceptable range - 0.4920 and 0.4132 for the crowd-sourced workers and the first round of expert annotations accordingly. 
However, the second round of expert annotations improved the inter-annotator agreement achieving a kappa coefficient of  0.7127.
For further insights, we computed standard deviations of annotator scores within the respective groups and present histograms of those statistics in Figure~\ref{fig:agreement_histogram_all}.
Plots of crowd-sourced annotations show strong similarities across all evaluated dimensions.
Such an effect could be caused by an insufficient distinction made by the annotators between the 4 scored axes, where the overall quality of a summary biased scores of the individual dimensions.
The histograms also show that while the second round of expert annotations lowered the standard deviation of scores and substantially increased inter-annotator agreement, relevance and coherence remained the most disagreed on dimensions between experts.
This could be attributed to the subjective nature of relevance and coherence as an evaluation dimensions~\citep{kryscinski-etal-2020-evaluating}.

\par
To assess the similarity of annotations between the crowd-sourced and expert annotators we averaged the assigned scores per example within the respective annotator groups and computed Pearson's correlation coefficient.
The statistic returned a value close to 0, indicating no correlation between expert and crowd-sourced judges.

\par
We also manually inspected the human annotations and present examples of annotated summaries, both generated and reference, as well as the differences in human judgments in Table~\ref{annotation_divergence_a}.
The first row shows a well written, comprehensive summary.
The high quality of the summary is reflected by top scores assigned by expert annotators, while being rated as average by crowd-sourced workers.
The second row shows a summary with ambiguous pronoun usage and factual inconsistencies.
The errors result in a decrease in coherence, consistency, and relevance scores in the expert annotations, but do not see a corresponding decrease in crowd-worker annotations.
The third row presents a factually correct summary that contains token and phrase repetitions.
The errors were caught by the expert annotators resulting in a low fluency score, while crowd-sourced annotators incorrectly classified them as issues with factual consistency.
These examples again illustrate the disparities in the understanding of evaluated dimensions between judges and underscore our observation above about the uniformity of crowd-sourced annotations; the crowd-sourced annotations tend to be similar across quality dimensions even when distinctions exist, which are captured in the expert annotations.
% annotation_divergence_b
%
\par
Results presented in this section highlight the difficulties of crowd-sourcing high-quality annotations and the necessity for protocols for improving human evaluation in text summarization. 

\subsection{Automatic Metrics}
Many automatic metrics have been proposed for evaluating both summarization and other text generation models.
However, the field lacks a comprehensive study that would offer a consistent side-by-side comparison of their performance.
We address this issue with the following experiments.

\par
In Table~\ref{correlations_11_system} we show Kendall's tau rank correlations between automatic metrics and human judgments calculated on a system-level following \citet{louis-nenkova-2013-automatically}.
The statistics were computed using the available expert annotations to avoid possible quality problems associated with crowd-sourced ratings, as highlighted in the previous subsection.
Automatic metrics were computed in a multi-reference setting, using the original reference summary included in the CNN/DailyMail dataset and 10 additional summaries coming from~\citet{kryscinski-etal-2020-evaluating}, and the length of model outputs was not constrained.
We report correlations without differentiating between abstractive and extractive models, as most metrics did not exhibit large differences in correlation when reported separately.
%
%For completeness, we include correlation tables for a setting with 1 and 6 reference summaries with a separation by model type in the Appendix.

%Despite the original intention of BLEU as a corpus-level metric we include its example-level correlations in this chart for completeness. 
%
%We also experimented with using a single references and with using the original and 5 additional references.
%
%We found that the correlations mostly increased with the additional references although they decreased slightly when going from 5 to 10 additional references.
%
%
%It is worth noting that only 4 of the 16 models analyzed for correlation were extractive. 

Correlation results show several trends.
We find that most metrics have the lowest correlation within the coherence dimension, where the correlation strength can be classified as weak or moderate.
This finding follows intuition as the majority of metrics rely on hard or soft subsequence alignments, which do not measure well the interdependence between consecutive sentences.
Low and moderate correlation scores were also found for the relevance dimension.
As discussed in the previous subsection, such trends could result from the inherent subjectiveness of the dimension and the difficulty of collecting consistent human annotations.
Model correlations increase considerably across the consistency and fluency dimensions.
While unexpected, the strong correlation with consistency could be attributed to the low abstractiveness of most neural models, which could increase the effectiveness of metrics using higher-order n-gram overlap, such as ROUGE-3 or Extractive Density. 
Referring back to the previous subsection, both of the mentioned dimensions achieved high inter-annotator agreement between expert judges which could also positively affect the correlation scores.
Additionally, the results show a substantially higher correlation between all evaluated dimensions and ROUGE scores computed for higher-order n-grams in comparison to ROUGE-L, which corroborates with findings of~\citet{rankel-etal-2013-decade}.

\par
To examine the dependencies between different metrics we computed Kendall's tau rank correlation coefficients, pairwise, between all metrics.
Results are presented as a correlation matrix in Figure~\ref{fig:pairwise_correlations_kendall}.
Following intuition, we observe a strong correlation between all metrics that compute, implicitly or explicitly, the lexical overlap between generated and reference summaries.
Metrics measuring the n-gram novelty and repetitiveness show a weak negative correlation with all ROUGE-related metrics.
Length as a feature is weakly correlated with most metrics apart from $S^3$, BLANC, and SuPERT which might suggest the mentioned metrics favor longer summaries.
Worth noting is also the weak correlation of reference-less SummaQA, BLANC, and SuPERT metrics with most other evaluated metrics.

\par
Results presented in this section highlight the evaluation dimensions that are not reliably covered by currently available metrics and pave the way for future work in model evaluation.

%% file: 6-model-evaluation.tex
\section{Model Re-evaluation}\label{sec:analysis}
\ModelBenchmarkTableA
%\ModelBenchmarkTableB

We now turn to an analysis of model scores across human evaluations and automatic metrics.
The evaluated models were released between 2017 and 2019, represent different approaches to summarization: abstractive, extractive, and hybrid, and their architectures reflect the trends in summarization research.
Although in many cases we obtained multiple variants of the same model, in the study we focus on the versions with the highest ROUGE-L scores.

\par
Table~\ref{models_human_judgments} contains the results of human evaluation across the four dimensions described in Section~\ref{sec:methodology-human-annotations}.
Scores for ground truth summaries are included as a point of reference.
We find that pretrained models such as Pegasus, BART, and T5 consistently performed best on most dimensions.
Notably, the mentioned models scored highest on consistency and fluency while obtaining lower scores for relevance and coherence.
Scores for extractive models highlight the known shortcomings of such approaches, which are lack of coherence of summaries and issues with selecting relevant content. 
%
%For the mentioned models consistency and fluency were not given a perfect score either due to the fluency of the chosen sentence or due to errors in the annotations, although generally, the scores were within reason.
%
Abstractive model ratings show an increasing trend with respect to the date of publication.
This is a promising result as it suggests that the quality of models is improving with time.
Worth noting is also the fact that reference summaries did not score well on consistency, coherence, and relevance.
Upon examination of the annotations, we found that the reference summaries often contained extraneous information, such as hyperlinks and click-bait descriptions of other articles.
As this information was not present in the source documents nor relevant for the summaries, the annotators interpreted it as hallucinations and assigned lower consistency and relevance scores.
Additionally, many reference summaries in the CNN/DailyMail dataset were constructed by naively concatenating bullet-point summaries into contiguous sequences. 
Such processing steps negatively affected the coherence of examples. 
Similar trends in human studies of reference summaries were reported by~\citet{stiennon-2020-gptsum}.
Examples of noisy reference summaries are shown in Table~\ref{annotation_divergence_b}.

\par
Table~\ref{model_scores_1} show scores for model outputs across all automatic evaluation metrics.
Parameters of metrics used in this study can be found in the evaluation toolkit repository listed in Section~\ref{sec:introduction}.
The results align with insights coming from the human evaluation of models.
We found that for most metrics, the highest scores were assigned to large models pretrained on vast quantities of data.
However, several metrics, such as $S^3$, SummaQA, SMS, CHRF, and METEOR tended to favor extractive models, assigning the highest scores to their outputs.

\par
Presented results provide a comprehensive perspective on the current state of the field and highlight directions for future modeling work.

%For the original ROUGE metrics, references and summaries were tokenized using the Stanford corenlp package.
%
%Otherwise, input was tokenized and stemmed according to initial implementations or left as is.
%
%A problem with evaluation on such a scale with models from different papers is the different tokenization strategies for the provided outputs.
%
%This results in metrics such as coverage, which should be 1.0 for extractive models, giving slightly different results.
% (Alex-DONE) BRYAN: "trends, and most importantly remained..." -> "trends, which remained consistent..."
%We performed experiments with several tokenization strategies for unifying the input, which resulted in slightly different scores but similar trends, which remained consistent with our experiments across models. 

%% file: 7-conclusions.tex
\section{Conclusions}\label{sec:conclusions}
We introduced SummEval,
a set of resources for summarization model and evaluation research that include:
a collection of summaries generated by recent summarization models on the CNN/DailyMail dataset, 
an extensible and unified toolkit for summarization model evaluation,
and a diverse collection of human annotations of model outputs collected from the crowd-source and expert annotators. 
Using the accumulated resources we re-evaluated a broad selection of current models and evaluation metrics in a consistent and comprehensive manner.
We hope that this work will prove to be a valuable resource for future research on text summarization evaluation and models. 
We also encourage the research community to join our efforts by contributing model outputs and extending the evaluation toolkit with new metrics.

%% file: 8-acknowledgements.tex
\section{Acknowledgements}
We thank all authors for sharing model outputs and Tony Wong for assistance with annotations.
% We thank all authors for sharing their model outputs and thus contributing to this work. %[ANON] and Tony Wong for assisting with annotations.

%% file: 9-appendix.tex
\section{Appendix}\label{sec:appendix}
% \subsection{Additional Correlation Tables}
% In the pages which follow we provide the remaining tables for correlations between automatic metrics and human judgments across four dimensions.
% \TableSystemAppendixAll
% \TableSystemAppendixHalf
% \TableSystemAppendixOne
% \clearpage
\paragraph{Data Collection}
The data collection interface used by both crowd-source and expert annotators is presented in Figure~\ref{fig:data_collection_ui}.
In the annotation process, judges were first asked to carefully read the content of the source article and next proceed to evaluating the associated summaries along four axes: \textit{relevance}, \textit{consistency}, \textit{fluency}, and \textit{coherence}.
\DataCollectionInterface